\documentclass{WileyMSP-template}

\usepackage{xcolor}
\usepackage{amssymb}
\usepackage{amsthm}
\usepackage{hyperref}
\usepackage{graphicx}
\usepackage{amsmath,amsfonts,amssymb}
\usepackage[normalem]{ulem}
\usepackage{soul}

\usepackage{xcolor}
\usepackage{amssymb}
\usepackage{amsthm}
\usepackage{hyperref}
\usepackage{graphicx}
\usepackage{amsmath,amsfonts,amssymb}
\usepackage[normalem]{ulem}
\usepackage{soul}
\usepackage{setspace}
\usepackage{lineno}
\usepackage{textcomp}
\usepackage{caption}
\usepackage{amsmath}           
\usepackage{etoolbox}          
\usepackage{setspace}
\usepackage[round, sort&compress, numbers]{natbib}
\usepackage{wrapfig}
\usepackage{enumitem}

\usepackage[utf8]{inputenc}
\usepackage[T1]{fontenc}
\usepackage{bm}

\usepackage[labelfont=bf,labelsep=period]{caption}

\begin{document}

\renewcommand{\baselinestretch}{1.5}
\setstretch{1.5}

\pagestyle{fancy}
\rhead{\includegraphics[width=1.6cm]{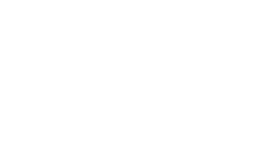}}
\title{Ultra-fast, programmable, and electronics-free soft robots enabled by snapping metacaps}
\maketitle

\author{Lishuai Jin}
\author{Yueying Yang}
\author{Bryan O. Torres Maldonado}
\author{Sebastian David Lee}
\author{Nadia Figueroa}
\author{Robert J. Full}
\author{Shu Yang*}


\begin{affiliations}
Dr. L. Jin, Y. Yang, Prof. S. Yang\\
Department of Materials Science and Engineering, University of Pennsylvania, 3231 Walnut Street, Philadelphia, PA 19104, USA\\
E-mail: shuyang@seas.upenn.edu\\
B. O. T. Maldonado, Prof. N. Figueroa \\
Department of Mechanical Engineering and Applied Mechanics, University of Pennsylvania, 220 S. 33rd Street, Philadelphia, PA 19104, USA\\
S. D. Lee\\
 Department of Mechanical Engineering, University of California at Berkeley, Berkeley, CA 94702, USA \\
Prof. R. J. Full\\
 Department of Integrative Biology, University of California at Berkeley, Berkeley, CA 94702, USA\\

\end{affiliations}


\noindent
\textbf{Abstract}

Soft robots have a myriad of potentials  because of their intrinsically compliant bodies, enabling safe interactions with humans and adaptability to unpredictable environments.
However, most of them have limited actuation speeds, require complex control systems, and lack sensing capabilities. To address these challenges, here we geometrically design a class of metacaps whose rich nonlinear mechanical behaviors can be harnessed to create soft robots with unprecedented functionalities. 
Specifically, we demonstrate a sensor-less metacap gripper that can grasp objects in 3.75 ms upon physical contact and a pneumatically actuated gripper with tunable actuation behaviors that have little dependence on the rate of input. Both grippers can be readily integrated into a robotic platform for practical applications.
Furthermore, we demonstrate that the metacap enables propelling of a swimming robot, exhibiting amplified swimming speed as well as untethered, electronics-free swimming with tunable speeds. 
Our metacaps provide new strategies to design the next-generation soft robots that require high transient output energy and are capable of autonomous and electronics-free maneuvering.


\clearpage

\section{Introduction}

Over the past decades, there have been considerable developments in soft robots that attempt to bridge the gap between conventional machines with high performance but rigid components and biological organisms with remarkable versatility and adaptability~\cite{cianchetti2018biomedical,wallin20183d,majidi2019soft,langowski2020soft,mccracken2020materials,rothemund2021hasel,li2022soft,rothemund2021shaping}.
The merits of soft robots are generally accomplished by deforming partial or all of the compliant robotic bodies via approaches such as pneumatic and hydraulic actuation~\cite{ilievski2011soft,whitesides2018soft,jin2021mechanical}, thermal stimulation~\cite{white2015programmable,wang2022multi}, solvent swelling~\cite{lee2010first,kim2021autonomous}, and application of magnetic and electric fields ~\cite{kim2018printing, sirbu2021electrostatic,diteesawat2021electro}. The deformation mechanisms can be roughly classified into two types: one deforms with a rate monotonically related to the input energy, but the performance is limited by the power and capacity of the input; the other exploits elastic structural instabilities to uncouple the input and output by gradually storing elastic energy before releasing it suddenly \cite{christianson2020cephalopod,pal2021exploiting,chi2022bistable,thuruthel2020bistable,tang2020leveraging,lin2021bioinspired,liu2022soft}. The latter is ideal for applications that require high-rate motion and fast energy release. 

Snap-through behaviors of spherical caps when tuned "inside-out" have been one of the most widely employed  mechanisms for achieving rapid responses in robots~\cite{gorissen2020inflatable,qiao2021bi}. However, the performance of these robotic systems is generally restricted by the intrinsic property of the caps. For instance, a simple spherical cap with uniform thickness and clamped border is typically monostable and extremely sensitive to imperfections when the thickness-to-radius ratio is small, leading to asymmetric deformation and limited energy release upon snapping (see \textbf{Figure 1}a and the corresponding pressure-volume curve in Figure 1d). While increasing the thickness-to-radius ratio can enhance the cap robustness, it also compromises the snapping behavior (see Figure 1b and the corresponding pressure-volume curve in Figure 1d).

Mechanical metamaterials, whose properties are determined by not only the constituent materials but also the architected geometries at the micro- and macro scales, offer a new paradigm to realize extraordinary snap-through behaviors and tunable stabilities by leveraging geometric constraints. These properties enable metamaterials with great potential in applications of "smart" purposes, such as energy absorption~\cite{shan2015multistable,yuan20193d}, biomedical devices~\cite{zanaty2019programmable,li2022miniature}, and soft robotics~\cite{chen2018harnessing,rothemund2018soft}. 

Here, we introduce the metamaterial concept to the spherical cap by designing a class of metacaps, which can realize different robotic functions enabled by the nonlinear yet tunable snapping behaviors.
The metacap comprises a spherical cap patterned with an array of ribs aligned in the circumferential and radial directions of the cap (Figure 1c). By rationally tailoring the dimensions and geometry of the ribs, we realize rich nonlinear mechanical properties of the caps, enabling a variety of soft robotic systems with unprecedented functionalities, including 
(i) a passive metacap gripper with mechanically embedded sensing capable of grasping objects in 3.75 ms upon contact; (ii) a pneumatically actuated metacap gripper with tunable actuation speeds that are independent of the input rate but readily tunable by changing the volume of the chamber connected to the gripper; (iii) a swimming robot whose speed is amplified by the metacap, which can be actuated untethered and electronics-free.

\section{Design of the metacap}

\begin{figure*}[ht!]
\centering
\includegraphics[width=1\linewidth]{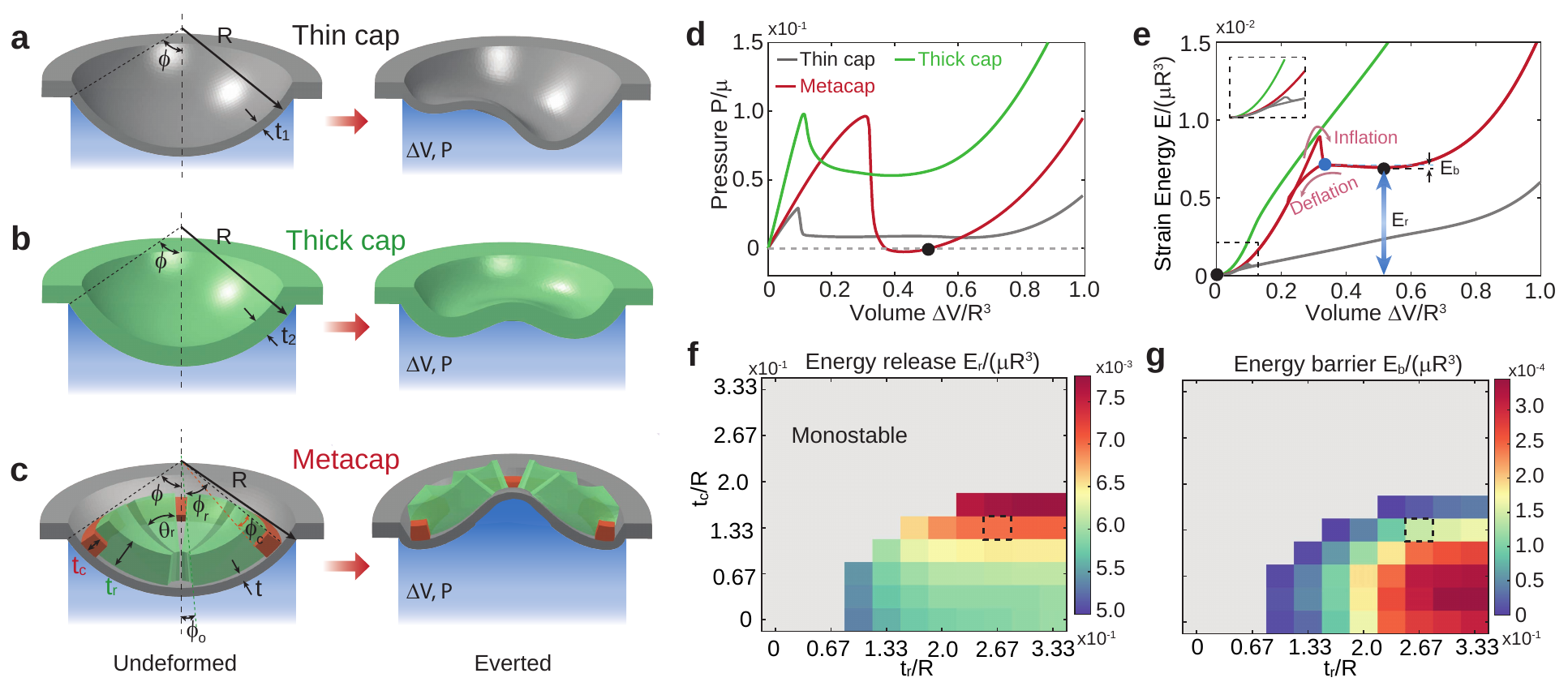}
\caption{ Design and characterization of the metacaps. {a}) A spherical cap with a low thickness-to-radius ratio exhibits an asymmetric snapping instability when inflated. b) A thick cap of the same radius is more robust showing a symmetric deformation mode but lacking of snapping. c) A metacap with architected structures, whose mechanical behaviors can be regulated by the geometries of the ribs. d) The  pressure-volume curves, normalized by initial shear modulus, $\mu$, and radius, $R$, of the thin cap ($t_1/R=0.075$, grey line), thick cap ($t_2/R=0.15$, green line) and metacap ($t/R =0.075$, $\phi_o = 5.0^\circ$, $\phi_c = 8.0^\circ$, $t_c/R=0.133$, $\phi_r = 47.85^\circ$, $\theta_r = 35.0^\circ$, $t_r/R=0.267$, red line). e) Evolution of the elastic energy as a function of the volume change for different caps. The inset shows the hysteresis of the thin cap upon inflation and deflation. f) Landscape of the energy release $E_r$ upon deflation as a function of $t_r$ and $t_c$. g) Landscape of the energy barrier $E_b$ upon deflation as a function of $t_r$ and $t_c$.} 
\label{fig:fig1}
\end{figure*}

To realize highly robust, bistable, and fast snap-through behaviors, we introduce an array of ribs to a simple spherical cap of radius, $R$, thickness, $t$, and polar angle, $\phi$. 
The array contains 8 radial ribs (green in color, with thickness, $t_r$, polar angle, $\phi_r$ and azimuthal angle, $\theta_r$, in Figure 1c) and one circumferential rib (red in color, with thickness, $t_c$, polar angle, $\phi_c$ ).
As the ribs affect the bending stiffness of the cap locally, the mechanical response of the metacaps can be significantly different from that of the spherical caps of uniform thickness.

First, we conduct finite element (FE) analyses to investigate the response of caps with variable geometries upon pressurization.
Figure 1d compares the pressure-volume curves of three representative caps, including the thin cap with $t_1/R=0.075$, the thick cap with $t_2/R=0.15$, and the metacap with $t/R =0.075$, $\phi_o = 5.0^\circ$, $\phi_c = 8.0^\circ$, $t_c/R=0.133$, $\phi_r = 47.85^\circ$, $\theta_r = 35.0^\circ$ and $t_r/R=0.267$. All have the same polar angle $\phi = 57.85^\circ$.
Upon inflation, a sharp pressure drop inside the metacap is observed when the applied volume is larger than the critical volume, which occurs at the peak pressure, indicating a more pronounced snap-through behavior than that of spherical caps of various thicknesses.  
Moreover, the metacap can achieve a new equilibrium state after snapping, which is highlighted by the black dot in Figure 1d.
In contrast, conventional spherical caps with clamped boundaries are monostable upon inflation, regardless of the thickness and radius~\cite{wagner2018robust}. Figure 1e plots the evolution of the strain energy  as a function of the volume change for the three caps. The inflation and deflation curves of the thick cap are identical, which increase monotonically as the applied volume increases. While the metacap and the thin cap show hysteresis between inflation and deflation curves due to the snap-through behavior. As expected, the thin cap has a monotonically decreasing curve upon deflation, implying that the thin cap is monostable. In contrast, the deflation curve of the metacap is nonconvex, indicating the bistability of the metacap, which is crucial for various robotic applications.

To shed light on how the geometry of the ribs affects the mechanical response of the metacap, we conduct a parametric study via FE simulations and report the landscape of the energy release, $E_r$, when the metacap deforms from the everted state to the undeformed state, and the energy barrier, $E_b$, which is required to trigger the transformation from the everted state to the undeformed state, in Figures 1f and 1g, respectively.
Throughout the study, we consider $R = 30~$mm, $\phi = 57.85^\circ$, $t/R =0.075$, $\phi_o = 5.0^\circ$, $\phi_c = 8.0^\circ$, $\phi_r = 47.85^\circ$ and $\theta_r = 35.0^\circ$ as fixed parameters and tune the response of the metacap by varying $t_c/R \in [0-0.33]$ and $t_r/R \in [0-0.33]$. Note that within this design space we can achieve metacaps with diverse nonlinear mechanical responses, and structures of other parameters can also be investigated using the same method.
From Figure 1f we find that the metacap is always monostable when $t_c/R\geq0.2$ or $t_r/R\leq0.067$ (the grey region in Figure 1f). For a given $t_r/R\geq0.1$, as the thickness of the circumferential ribs $t_c$ increases, $E_r$ monotonically increases so that the released energy becomes larger and larger. While the effect of $t_r$ on $E_r$ is less pronounced, characterized by little variation of $E_r$ when $t_c$ is fixed. In contrast, both $t_r$ and $t_c$ have a significant impact on the energy barrier $E_b$; a smaller $E_b$ can be achieved around the border between monostable and bistable regions, indicating that the metacaps from that region can be easily triggered to transform from the everted state to the undeformed state.
 
\section{Ultra-fast, passive metacap grippers}

\begin{figure*}[ht!]
\centering
\includegraphics[width=1\linewidth]{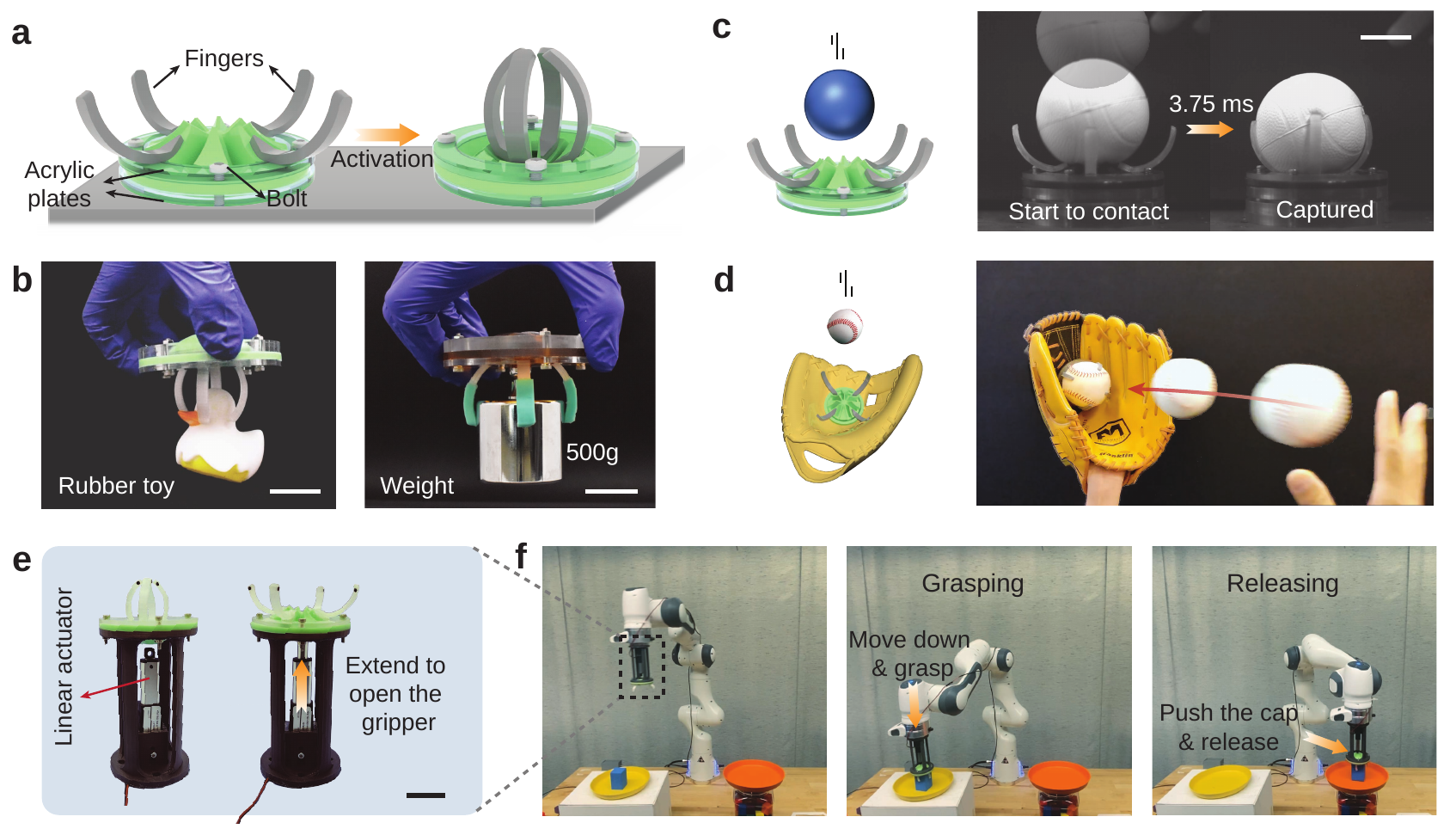}
\caption{ Passive metacap grippers.  a) Schematics of the passive gripper that closes upon applying a certain force at the center of the cap. The cap is fixed by two acrylic plates through four bolts, and four 3D printed fingers (ABS) are glued on the ribs of the cap to catch the objects. b) The grippers can grasp objects of different shapes, moduli, and weights. c) The gripper is capable of realizing highly dynamic grasping tasks. A ball hits at the center of the cap with a speed of around 5.8 m/s and is caught by the gripper in 3.75 ms. d) Integration of the gripper to a baseball glove facilitates the grasping of a baseball. e, f) The metacap gripper is integrated into a robotic arm (Franka Emika Panda) and controlled by a linear actuator (USLICCX LA-T8) that opens it. Closing is fully passive. Scale bars, 30 mm.}\label{fig:fig2}
\end{figure*}

To test the unique mechanical behaviors of the metacap, we design a passive gripper to grasp objects when a certain contact force is applied to the center of the metacap. The passive gripper comprises a bistable metacap, four fingers (3D printed from Acrylonitrile Butadiene Styrene, ABS) evenly glued on the radial ribs of the cap, and two acrylic plates fixed by four bolts to clamp the border of the metacap (see \textbf{Figure 2}a and Section S1.1 for fabrication details). By varying the geometry and material of the cap, we can adjust the stiffness of the gripper to grasp objects of different shapes, moduli, and weights. Figure 2b shows experimental snapshots of two grippers made of Elite Double 32 (with green color and initial shear modulus $\mu =0.36~ $MPa) and Flex 80 (with brown color and initial shear modulus $\mu =1.55~ $MPa) elastomers with the geometry highlighted in the dotted box in Figures 1f and 1g, grasping a rubber toy and a 500 g weight ($\sim$10x their self-weight), respectively. Note that the fingers of the Flex 80 gripper are coated with a thin layer of elastomer (Elite Double 22) to enhance friction for stable grasping. Although the fingers are printed from rigid ABS, the stiffness of the gripper and its response time is determined by the metacap instead of the fingers. More grasping tests on different objects can be found in Figure S18.

As the bistable metacap releases a large amount of elastic energy when the gripper deforms from the open to the closed state, the gripper is able to react rapidly and perform highly dynamic grasping tasks. Figure 2c shows experimental snapshots of the metacap gripper made of Flex 80, grasping a stress ball in 3.75 ms after it hits the center of the metacap at a speed of $\sim$5.8 m/s. 
When the gripper is integrated into a baseball glove, it facilitates catching without the need for fingers(see snapshots in Figure 2d), which will help with hand movement practice. This feature would be especially beneficial for people with hand disabilities to play baseball (see Movie S3 and fabrication of the glove with the gripper in Figure S3). 
Furthermore, the rapid response of the metacap gripper shows great potentials for locomotion tasks on thin, cantilevered beams. A real life example is that of squirrels'  landing behavior when they leap from branch to branch (Figure S12). Squirrels are known for their acrobatic maneuvers and capability of leaping through complex tree canopies to travel and avoid predators \cite{hunt2021acrobatic}, which makes them  exemplary models to emulate for high-mobility robots. Our metacap gripper can mimic the landing behavior of squirrels' paws and facilitate the design of biomimetic robots with high agility~\cite{roderick2021bird}.

The simplicity of our metacap gripper makes it easy to integrate into an existing robotic platform for industrial applications. Towards this end, we clamp the gripper onto a 3D printed frame and mount a linear actuator (USLICCX LA-T8) at the center of the frame (Figure 2e) to open the gripper by pushing the cap using the shaft of the linear actuator. Since the gripper is passive, it can grasp the object automatically when in contact with the object above a certain force. As seen in Figure 2f, the objects can be grasped by the gripper when the robotic arm moves down and be released at the target position by pushing the metacap using the linear actuator (Movie S5).

\section{Pneumatically actuated grippers with tunable actuation speeds}

\begin{figure*}[b!]
\centering
\includegraphics[width=1.0\linewidth]{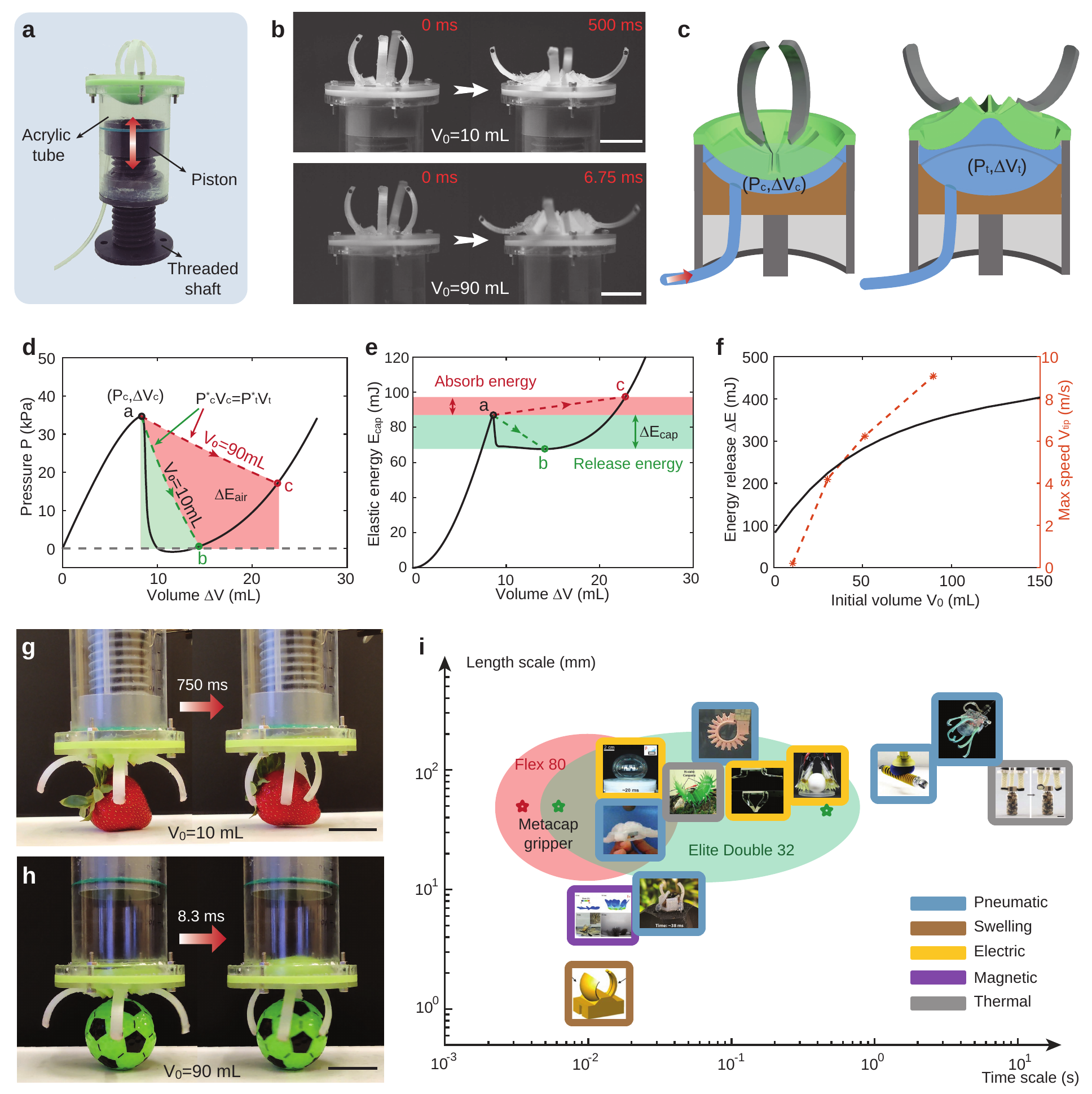}
\captionsetup{labelformat=empty}
\caption{
  }\label{fig:fig3}
\end{figure*}
\addtocounter{figure}{-1}

\begin{figure*}[t!]
\caption{
Pneumatically actuated grippers with tunable speeds. a) An experimental snapshot of the actuator. The metacap gripper is clamped to an acrylic chamber  whose volume is tunable by changing the position of the 3D-printed piston.  b) Experimental snapshots  of the opening process of grippers with different initial volumes. The one with $V_0 = 10$ mL takes 500 ms to deform from the closed state to the open state, while the one with $V_0 = 90$ mL needs 6.75 ms to snap open, indicating that actuation speeds can be tuned by the initial volume of the chamber. c) Schematic of the pneumatically-actuated gripper before and after snapping. $P_c$ and $\Delta V_c$ are the critical pressure and volume change inside the chamber before the metacap snaps. $P_t$ and $\Delta V_t$ represent the pressure and volume change after the metacap snaps. d)  The energy release from compressed air during the snapping process can be calculated from the pressure-volume curve of the metacap via Boyle's law. 
e) Evolution of the strain energy of the metacap as a function of volume change. When the initial volume $V_0 = 10$ mL, the metacap releases 19.3 mJ during the snapping, while  absorbing 10.3 mJ when $V_0 = 90$ mL.
f) Evolution of the total energy release calculated from Equation~(\ref{eq1}) and the maximum speed of the gripper as a function of the initial volume $V_0$. g) Experimental snapshots of the metacap gripper with $V_0 =10$ mL, taking 750 ms to grasp a strawberry gently.  h) Experimental snapshots of the metacap gripper with $V_0 =90$ mL,  taking 8.3 ms to grab a plastic ball. i) Comparison of the actuation speed and dimensions from our metacap grippers vs. those from the  soft actuators reported in the literature based on pneumatic actuation \cite{brown2010universal,thuruthel2020bistable,mosadegh2014pneumatic,sinatra2019ultragentle,bas2021ultrafast}, swelling~\cite{lee2010first}, application of electric~\cite{acome2018hydraulically,baumgartner2020lesson,wang2019soft} and magnetic \cite{wang2020untethered} fields, and thermal activation \cite{kim2010towards,he2019electrically}. Our grippers exhibit the fastest actuation speed while still being tunable for multipurpose applications. Scale bars, 30 mm.}
\end{figure*}

When the passive metacap gripper grasps objects rapidly, it also inflicts high impacts onto the objects, causing undesirable damage to soft and delicate objects such as fruits and eggs.  
To extend the capability and adaptiveness of our grippers, we design a pneumatic actuation mechanism  (\textbf{Figure 3}a) where the actuation speed can be fine-tuned to allow for not only the rapid grasping for highly dynamic tasks but also the gentle manipulation of delicate objects. We clamp the metacap gripper to a chamber formed by an acrylic tube and a 3D-printed piston, whose volume  can be adjusted by changing the position of the piston via twisting the translational screw that's connected to the piston (see Section S1.4 about the fabrication of the pneumatically actuated gripper). By leveraging the compressibility of air and the bistability of the metacap, 
we achieve gentle actuation when the initial volume of the chamber $V_0$ is small (e.g. 10 mL in Figure 3b) while realizing rapid grasping when $V_0$ is large (e.g. 90 mL in Figure 3b). 
The metacap is made of Elite Double 32 with the geometry highlighted in Figures 1f and 1g. It takes 500 ms for the gripper with $V_0 = 10$ mL to snap open when inflated using a diaphragm  pump (JSB1523006, TSC, China ), but 6.6 ms to snap open when $V_0 = 90$ mL with the same input (Movie S6). 

To quantitatively understand the effect of $V_0$ on the actuation speed of the gripper under a constant pneumatic input, we develop a simple model to investigate the energy release during the opening and closing of the gripper. We denote the critical pressure and the critical volume change that the cap snaps at as $P_c$ and $\Delta V_c$, respectively, and the pressure and the volume change after the cap snaps as $P_t$ and $\Delta V_t$, respectively (Figure 3c). The actuation speed is determined by the amount of energy release during the snapping. We ignore the energy from the input because the snapping is fast and the input flow rate is relatively slow. Therefore, the total energy release contains the elastic energy change from the metacap $\Delta E_{cap}$ and the compressed air $\Delta E_{air}$ as
\begin{align}\label{eq1}
    \Delta E=\Delta E_{cap}+\Delta E_{air}.
\end{align}
Since air is compressible, both $\Delta E_{cap}$ and $\Delta E_{air}$ are affected by $V_0$. Here we assume that the snapping is an isothermal process and the status of the cap after snapping can be determined via Boyle's law 
\begin{align}\label{eq2}
    P_c^*V_c=P_t^*V_t,
\end{align}
where $P_x^*=P_0+P_x$ (with $x = c$ or $t$) is the absolute pressure with $P_0 =101.3$ kPa as the standard atmospheric pressure, $V_x=V_0+\Delta V_x$ is the total volume inside the chamber, and the subscripts $c$ and $t$ represent the states before and after the cap snaps, respectively.

According to Equation~(\ref{eq2}), one can predict the state of the snapped metacap from the initial state of the chamber. For instance, the metacap snaps from state $a$ to state $b$ (Figure 3d) when $V_0 = 10~ $mL, and the compressed air releases 97.6 mJ during the snapping process, which can be calculated by computing the green area shown in Figure 3d. When $V_0 = 90~ $mL, the metacap snaps from state $a$ to state $c$, releasing 370.2 mJ from the compressed air, equal to the summed area of green and red regions. Moreover, the energy release from the metacap can be calculated from the elastic energy-volume change curve shown in Figure 3e, that is 19.3 mJ and -10.3 mJ for $V_0 = 10$ mL and 90 mL, respectively. The negative sign indicates that the metacap will absorb energy if $V_0$ is large enough. In light of Equation~(\ref{eq1}), the total energy release for $V_0 = 10~$mL and 90 mL is 116.9 mJ and 359.9 mJ, respectively. As expected, more energy can be released from larger $V_0$, which explains the faster actuation speed observed in Figure 3b when $V_0$ = 90 mL.

In Figure 3f we report the evolution of total energy release as a function of $V_0$, showing a monotonic relationship. Likewise, the maximum speed of the fingertips of the gripper show a similar trend with $V_0$ as predicted by our model, i.e. larger initial volume results in higher snapping speed. To demonstrate the capability of adjusting the actuation speed for diverse objects, we integrate our gripper into a robotic arm and grasp soft objects (i.e. a strawberry) using a gentle actuation (750 ms, $V_0 = 10$ mL) to avoid potential damage (Figure 3g),  while using fast actuation speeds (8.3 ms, $V_0 = 90 $ mL) to grasp strong and rigid objects for highly dynamic tasks (Figure 3h). Note that the closing process of the gripper is slower than the opening with the same $V_0$ and input flow rate because less energy is released during the closing process (Figure S19).

The motion of soft robots is mostly affected by the rate of deformation and elastic recovery of their compliant bodies - conventional soft actuators are either very slow when designed with large dimensions \cite{mosadegh2014pneumatic,sinatra2019ultragentle}, or very small to acquire high actuation speed \cite{bas2021ultrafast,lee2010first}.  Our passive grippers and pneumatically actuated grippers harness the extraordinary snap-through behavior of metacaps and the compressibility of air, outperforming the conventional soft grippers in terms of actuation speed and tunability.
In Figure 3i we compare the actuation speed and dimensions between our grippers with a few representative  soft actuators reported in the literature via different actuation mechanisms. Our grippers exhibit the fastest actuation speed compared with others of actuation time ranging from 12 ms to 20 s. Furthermore, our metacap grippers can benefit from the tunable actuation speeds, spanning a wide range of time scales (3.75 ms to 750 ms) for on-demand actuation, facilitating multipurpose grasping tasks with the same design. Importantly, the simple geometry and  fabrication process make our metacap grippers easy to scale up or scale down for  applications at different length scales.

\section{Swimming robots}

\begin{figure*}[ht!]
\centering
\includegraphics[width=1\linewidth]{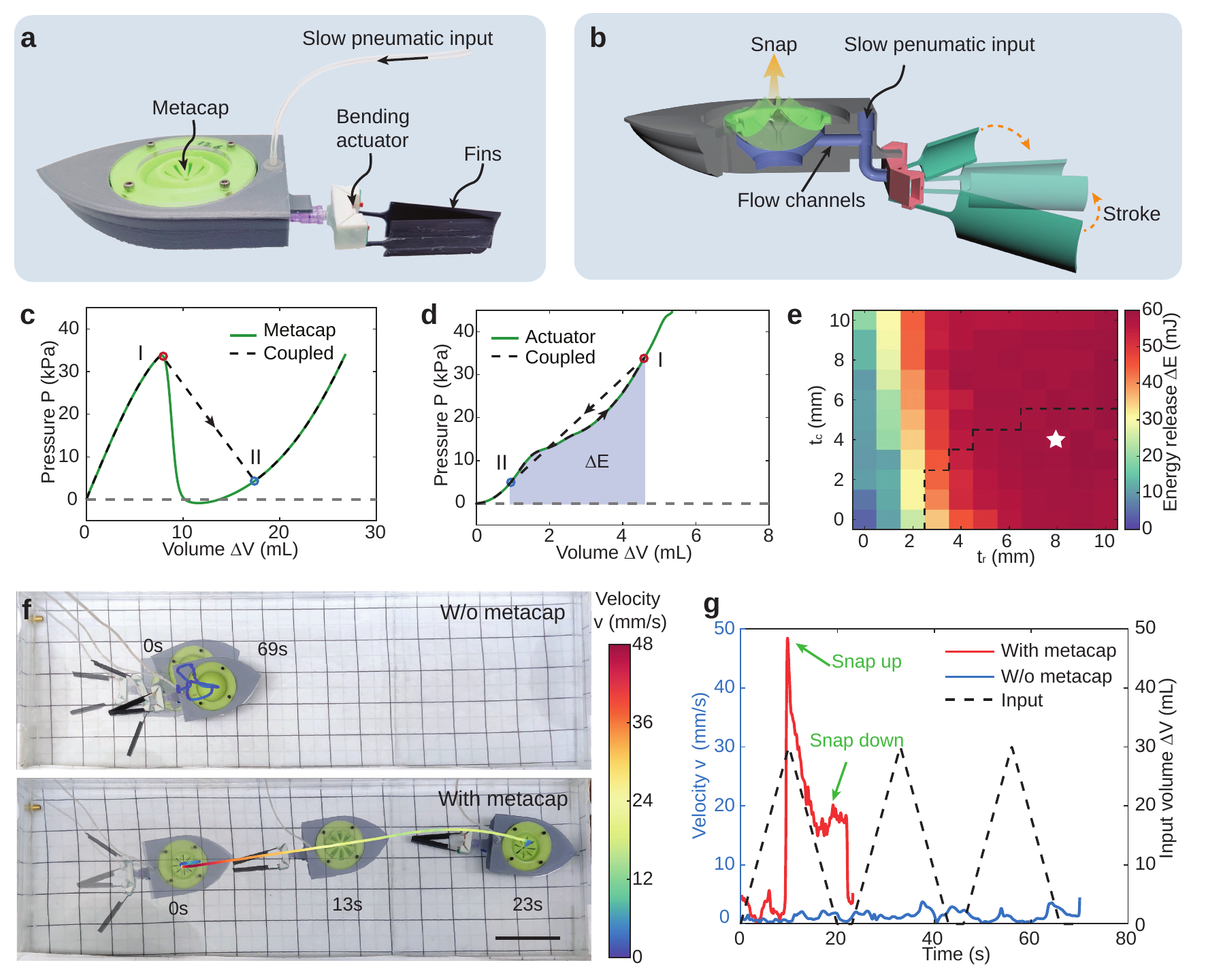}
\caption{ The metacaps enable swimming robots with amplified swimming speed. a) An experimental snapshot of the swimming robot comprising a 3D-printed body, a metacap, and a bending actuator that controls the movement of the fins.  b) Schematic of the mechanism of the swimming robot.  c, d) Pressure-volume curves of (c) the metacap and (d) the bending actuator when inflated with incompressible fluid (continuous lines); and the responses of the metacap and the bending actuator when they are coupled through the flow channels and inflated with air (dashed lines). e) The landscape of energy release from the bending actuator for each stroke as a function of the thickness of $t_c$ and $t_r$. f) Experimental snapshots of the swimming robots with and without the metacap, both of them are supplied with the same pneumatic input. Scale bar, 10 cm. g) Comparison  of the instantaneous velocity between these two robots together with the profile of the input volume. }\label{fig:fig5}
\end{figure*}

Besides gripping, our metacaps can be utilized to amplify the actuation speed of other actuators for  applications that require high transient output power. We exemplify this by incorporating the metacap with a bending actuator (see details in Figures S5, S6, and S16) and design a robot capable of swimming rapidly with a relatively slow pneumatic input (\textbf{Figure 4}a). We connect the metacap and the bending actuator via a 3D-printed robotic body (ABS) embedded with flow channels for air exchange (Figure 4), providing buoyancy for the robot to float in water. 
Two fins are mounted at the ends of the bending actuator to generate propulsion for swimming. When supplied with a relatively slow pneumatic input, the bending actuator gradually deforms and opens the fins, while the metacap will not snap until the pressure inside the chamber reaches the critical pressure $P_c$. Once the cap snaps, the pressure inside the chamber decreases immediately,  and the bending actuator quickly deforms back, driving the fins to generate a large propulsion for swimming. In contrast, if the metacap is replaced by an elastic membrane that cannot snap, the bending actuator deforms monotonically to the input flow rate, and the fins are not able to generate large enough propulsion for swimming. 

To maximize the propulsive energy from the bending actuator, we use a simple model to quantify the energy release of the bending actuator during the metacap snapping. Specifically, we employ pressure-volume curves of the  metacap and bending actuator (continuous lines in Figures 4c and 4d) to predict the behavior of the entire system when the metacap and the bending actuator are mechanically coupled through the flow channels (dashed lines in Figures 4c and 4d). Upon applying a slow pneumatic input, the metacap and the bending actuator first gradually deform to point I in Figures 4c and 4d. After snapping, the metacap and the bending actuator  deform immediately to configurations that correspond to point II under the following constraints
\begin{align}\label{eq3}
    &P_{cap}(\Delta V_{cap})=P_{actuator}(\Delta V_{actuator}),\\
    &\Delta V_{cap}+\Delta V_{actuator}=\Delta V,
\end{align}
where $P_{cap}(\Delta V_{cap})$ and $P_{actuator}(\Delta V_{actuator})$ represent the pressure-volume relationships of the metacap and the bending actuator. $\Delta V_{cap}$ and $\Delta V_{actuator}$ are the volume change induced by the metacap and actuator upon pressurization, respectively. With this model, one can predict the energy release $\Delta E$ from the bending actuator by calculating the area of the blue region shown in Figure 4d. To identify the best design of the metacap that makes the robot swim rapidly, we systematically explore the parameter space of the metacap, and the energy release landscape of the bending actuator is shown in Figure 4e. The effect of $t_r$ on the  energy release is more pronounced than that of $t_c$; when $t_r$ is large enough ($t_r \geq$ 6 mm) the  energy release almost stabilizes to a constant. However, the parameters above the dashed lines lead to undesired monostable metacaps, because the residual pressure inside the chamber after metacap snaps prevents the bending actuator from closing completely, and the open fins give rise to large resistance during swimming.

For proof-of-principle, we test two robots with and without the metacap in a water tank (30 cm in width and 90 cm in length, Figure 4f). The metacap has $t_r = 8 $ mm and $t_c =4 $ mm (highlighted in Figure 4e). By supplying with the same input (see the input profile in Figure 4g), the swimming robot with a non-snapping membrane barely swims forward, while the one with the metacap can swim forward with a much faster speed after the cap snaps (Movie S7). The lines in the figures represent the trajectories of the robots, and the colors of the lines indicate the instantaneous velocity of the robots. In Figure 4g, we compare the instantaneous velocity of these two robots, and it is clear that the robot with a metacap has a much faster speed (more than 10 times larger in maximum speed) characterized by two peaks as a result of the metacap snapping up and down.

\section{Untethered, electronics-free robots enabled by oscillating valves}

\begin{figure*}[ht!]
\centering
\includegraphics[width=1.0\linewidth]{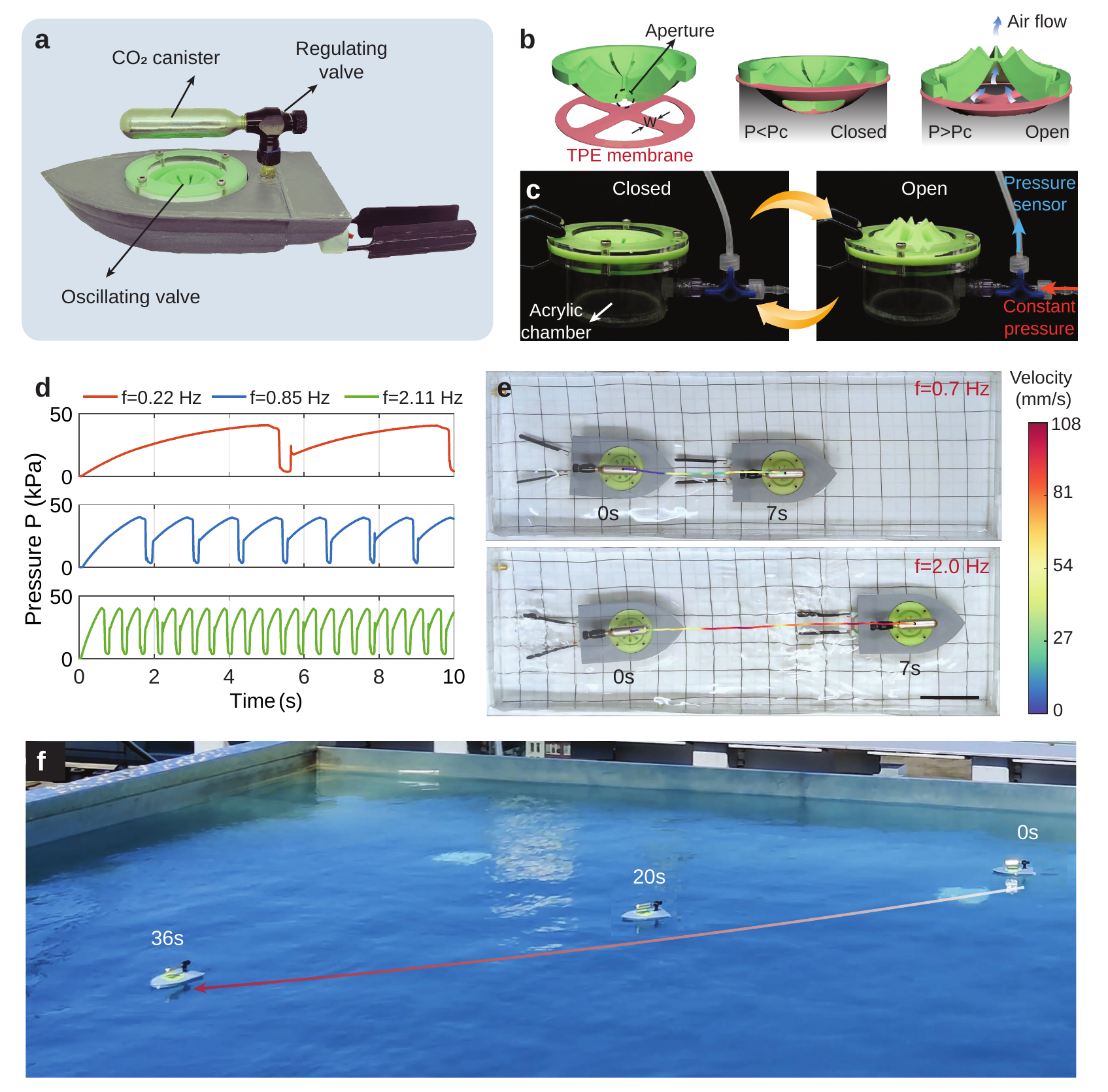}
\caption{Oscillating valves enable untethered and electronics-free swimming robots.  a) An experimental snapshot of the untethered and electronics-free swimming robot. b) Design and working mechanism of the oscillating valve. The valve comprises a monostable metacap with an aperture at the center of the cap and a TPE membrane.  c) Experimental setup for characterizing the oscillating valve.  d) Various pressure profiles are obtained from varying the input flow rate of the valve. e) Experimental comparison of the swimming speed for  robots with different stroke frequencies. Scale bar, 10 cm. f) Experimental snapshots of the robot tested in a swimming pool.  }\label{fig:fig6}
\end{figure*}

Although our metacap can accelerate the speed of the swimming robot significantly, the robot is controlled by a tethered electronic pump. To make the robot electronics-free and untethered, we introduce an oscillating valve actuated by a CO$_2$ canister (\textbf{Figure 5}a).
The valve consists of a monostable metacap with an aperture (3 mm in diameter) at the center of the cap, which is covered by a thermoplastic elastomer (TPE) membrane (Figures 5b and S7).
When the pressure inside the chamber $P$ is smaller than $P_c$, the aperture remains sealed by the membrane. However, when $P$ is larger than $P_c$, the cap snaps, opening the valve. Thus, the air inside the chamber flows out through the aperture, leading to a dramatic pressure decrease inside the chamber. Since the metacap is monostable, it will snap back when $P$ is lower than a certain value ($\sim$ 3.5 kPa) and the aperture will be sealed by the membrane again. Hence, the valve exhibits an oscillating response when it is supplied with a constant pneumatic input.

To characterize the oscillating response of the valve, we mount a monostable metacap (with $t_c = 8 $ mm, $t_r = 8 $ mm, $\phi_c = 12 ^\circ$) and a TPE membrane (with $w$ = 7 mm) to an acrylic chamber with a volume of 100 mL (Figure 5c) and monitor the evolution of the pressure inside the chamber using a pressure sensor (ELVH-015D, All Sensors, Figure S14). By supplying the chamber with a constant pressure input, the metacap snaps up and down periodically, resulting in an oscillatory pressure profile inside the chamber, where the frequency of the valve can be readily tuned by varying the flow rate of the input. In Figure 5d, we show three pressure profiles of the same valve when supplied with a constant pressure ($\sim50$ kPa) at low (the red curve), medium (the blue curve), and high (the green curve) rates, and the oscillatory frequency of the valve varies from 0.22 to 2.11 Hz. The maximum and minimum oscillating frequencies for each valve are determined by the geometry of the metacap and the membrane, as well as the cavity of the chamber. When fixing the geometry of the metacap and the chamber, but varying the width (w) of the membrane (Figure 5b), we obtain a wide range of oscillating frequencies $f \in [0.12, 4.25]$ when $w \in [3, 11]$ mm (Figure S15 and Movie S8).
Although a few oscillating valves have been reported ~\cite{preston2019soft,lee2022buckling,van2022fluidic,Decker2022}, ours outperforms the existing systems in three aspects: (i) it does not require a complex circuit to control oscillation; (ii) it exhibits a sharp pressure drop in each cycle, which is beneficial for rapidly responsive robots;  (iii) a wide range of oscillating frequencies can be achieved by simply tuning the input flow rate.

With the oscillating valve, our swimming robot can not only be untethered and  electronics-free, but also have tunable swimming speed  by  varying the input flow rate from the CO$_2$ canister. Figure 5e shows that the robot can achieve stroke frequencies ranging from  0.7 Hz to 2.0 Hz, capable of swimming 29 cm and 50 cm in 7 s, respectively. The lines in the figures represent the trajectories of the robots, and the colors of the lines indicate the instantaneous velocity of the robots.
Furthermore, our swimming robot is highly efficient as it can work for more than 1 min at $\sim$ 1 Hz when supplied with a 12 g CO$_2$ canister, and it can cross the diagonal of a swimming pool ($\sim$6x6 $m^2$) in 40 s (Figure 5f and Movie S9).

\section{Conclusions}

We have developed a class of metacaps composed of a simple spherical cap and an array of ribs. Through finite element simulations, numerical modeling, and experimental demonstrations, we investigated the geometric effect of the metacap on its mechanical responses, the interaction between metacaps and compressible air, and the coupling between metacaps and conventional bending actuators. The rich nonlinear mechanical responses of the metacaps and the remarkable interactions between metacaps, air, and conventional actuators enable several robotic systems with ultrafast yet tunable actuation speed without the need for external sensors  and complex control systems.

The passive grippers are capable of grasping diverse objects automatically upon application of a certain contact force. The fastest actuation speed (3.75 ms for grasping) is achieved in comparison to other soft grippers reported in the literature. We show that the passive grippers have great potential in sports equipment and industrial applications due to their remarkable adaptability and maneuverability, simple geometry, low cost, and ease of fabrication. The pneumatically controlled grippers uncouple the actuation speed from the input and realize tunable actuation speed by adjusting the initial volume of the chamber. 

Lastly, we demonstrated that our metacaps can be used to pneumatically regulate swimming robots for high transient energy output, exhibiting remarkably higher efficiency than those without the metacap. In turn, our study provides  valuable insights into swimming and underwater robots for oceanic exploration. By varying the input flow rate, our oscillating valves exhibit a wide range of oscillating frequencies, providing a new design solution for untethered and electronics-free pneumatic robots, which are beneficial for  applications that are sensitive to spark ignition (e.g. some rescue circumstances may be full of explosive gas). 

The  metacap has allowed us to infiltrate into the property space that was previously inaccessible  for conventional metamaterials and has uncovered novel soft robots with unprecedented performances which are not achievable with simple spherical caps of uniform thickness.
In particular, our metacaps hold promise for numerous applications that require high transient output energy in biomedical engineering and robotic systems, ranging from ventricular assist devices  \cite{gonccalves2003dynamic,timms2011review} and soft mechanotherapy devices \cite{preston2019soft} to locomotive and jumping robots \cite{drotman2021electronics,Hong2022}.  We envision that the concept of metacap demonstrated here would enrich the  design palette of soft robots that are ultra-fast, programmable, autonomous, and electronics-free.

\medskip
\noindent
\textbf{Supporting Information} \\Supporting Information is available from the Wiley Online Library or from the author.\\

\textbf{Acknowledgements}\\ We gratefully appreciate Dr. Zhongdong Jiao for his help on manufacturing the pump system. We acknowledge the usage of the Instron supported by the Department of Materials Science and Engineering Departmental Laboratory at the University of Pennsylvania.
This research is supported by Army Research Offices (ARO) through the MURI program, ARO \# W911-NF-1810327.  \\

\textbf{Author contributions}\\
 S.Y. and L.J. conceived the research idea. L.J. designed the metacap, built the model, conducted the finite element simulations, and proposed the robotic demonstrations. L.J. and Y.Y. fabricated the prototypes. L.J., Y.Y., B.O.T.M., and N.F. conducted the experimental testing. S.D.L. and R.J.F. tested the squirrels' behaviors. S.Y. supervised the research. L.J. and S.Y. wrote the manuscript. All authors contributed to the editing of the manuscript.\\
 
\textbf{Competing interests } \\The metacap design and robotic demonstrations presented in this work have been filed under a provisional patent application.

\textbf{Data Availability Statement }\\
All data needed to evaluate the conclusions in this study are present in the paper and/or the Supplementary Information.


\newpage
\begin{figure}
\centering
 \textbf{Table of Contents}\\
 \medskip
   \includegraphics [width=0.7\linewidth]{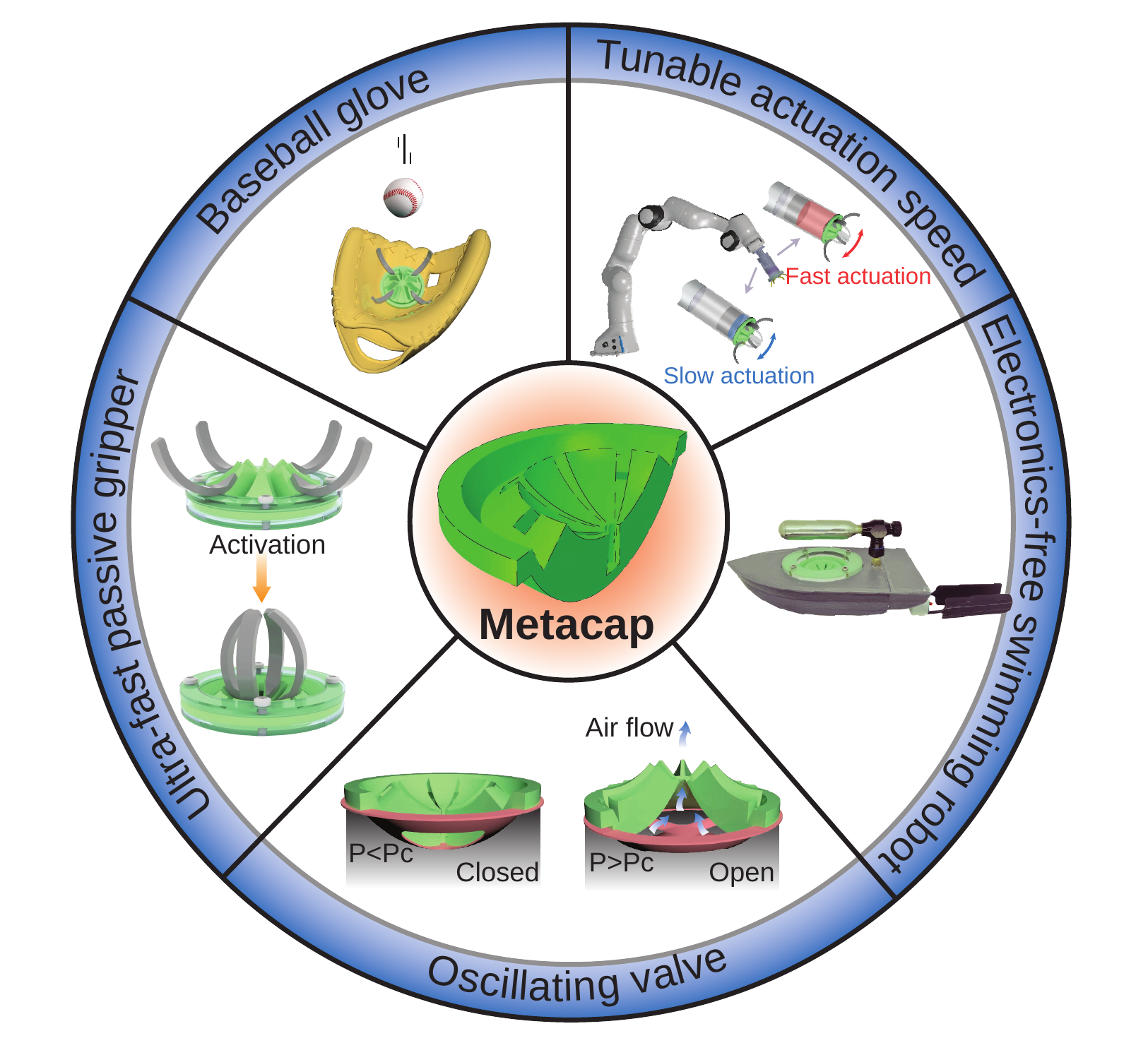}
   \medskip
   \caption*{
This work presents a new class of metacaps whose extraordinary nonlinear properties have been harnessed to enable several robotic systems with ultra-fast yet tunable actuation speed without the need for external sensors or complex control systems.  The metacap design strategies pave the way for next-generation soft robots that are ultra-fast, programmable, autonomous, and electronics-free.  }

 \end{figure}

\end{document}